\documentclass[runningheads]{llncs}
\usepackage{graphicx}
\usepackage{booktabs} 
\usepackage{multirow}  
\usepackage[ruled,vlined]{algorithm2e}
\SetKwInput{KwInput}{Input}

\usepackage{tikz}
\usepackage{comment,verbatim}
\usepackage{amsmath,amssymb} 
\usepackage{color}
\usepackage{cite}

\usepackage[accsupp]{axessibility}  
\usepackage[table]{xcolor}
\usepackage{colortbl}
\usepackage{multirow}
\usepackage{booktabs}

\definecolor{PriorRow}{RGB}{255,241,224}   
\definecolor{AblRow}{RGB}{230,244,234}      
\definecolor{OursRow}{RGB}{219,237,250}   

\newcommand\blfootnote[1]{%
  \begingroup
  \renewcommand\thefootnote{}\footnote{#1}%
  \addtocounter{footnote}{-1}%
  \endgroup
}

\begin{document}
\pagestyle{headings}
\mainmatter

\title{SegWave: Wavelet-Driven Segmentation of Tampered Regions} 

\titlerunning{SegWave: Wavelet-Driven Segmentation of Tampered Regions} 

\author{Siddhi Pravin Lipare \and
Vishesh Kumar \and
Akshay Agarwal}

\authorrunning{Lipare et al.} 


\author{Siddhi Pravin Lipare\inst{1}$^{\dagger}$ \and
Vishesh Kumar\inst{2} \and
Akshay Agarwal\inst{2}}
\authorrunning{Lipare et al.}
\institute{IIIT-Hyderabad, India\\
\email{siddhi.l@research.iiit.ac.in}
\and
Trustworthy BiometraVision Lab, IISER-Bhopal, India\\
\email{\{vishesh22, akagarwal\}@iiserb.ac.in}}

\maketitle
\blfootnote{$^{\dagger}$ Work done while at IISER Bhopal.}
\begin{abstract}

Verifying image authenticity is increasingly difficult, posing serious risks across journalism, law enforcement, and political domains. Most existing forensic methods rely on high-level visual artifacts and treat frame detection as a simple binary task. To address this, we propose \textbf{SegWave}, a hybrid framework that jointly leverages spatial and frequency-domain cues for image tampering detection. SegWave integrates a transformer-based architecture with the Discrete Wavelet Transform (DWT) to capture localized, multi-scale frequency inconsistencies indicative of manipulation. To further improve localization effectiveness, we introduce an \textbf{Adaptive Sub-band Attention module} (ASA) that dynamically highlights the informative high-frequency wavelet components. Extensive experiments on multiple benchmark datasets demonstrate that SegWave consistently outperforms state-of-the-art tampering detection methods in challenging evaluation settings.

\end{abstract}

\section{Introduction}
\label{sec:intro}

Verifying whether an image is authentic has become central to the modern information ecosystem - from journalism and legal evidence to public discourse on social media. With the rapid progress of generative AI~\cite{9887823,realistic,ostf}, images can now be synthesized and edited with a realism that is difficult to spot by eye. The same tools that enable creative work let malicious actors fabricate evidence, alter surveillance footage, or spread convincing visual misinformation. As manipulations become more sophisticated, the demand for tampering detection algorithms that can keep pace grows just as quickly.

Consider an image circulating online that shows a public figure at an event they never attended. The face may be spliced from one photograph, the background lifted from another, and a small object inserted by a diffusion model - three distinct manipulations composited into a single frame. For a journalist verifying the image, an analyst preparing it for court, or an automated moderation system, a verdict of simply ``tampered'' is of little practical use. What these users actually need is to know \textit{where} the image is altered and \textit{how many distinct sources} are stitched together, so that the manipulation can be traced, attributed, and explained.

\begin{figure*}[t]
    \centering
    \includegraphics[width=1\linewidth]{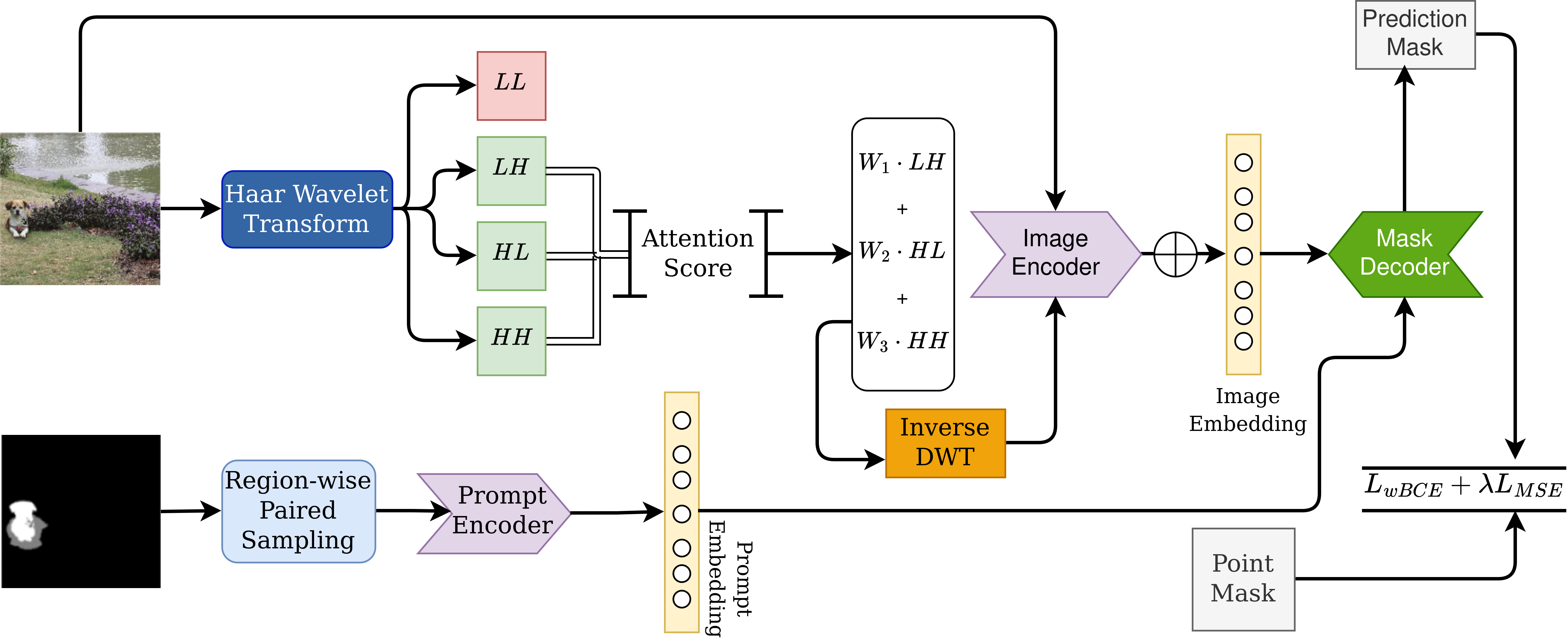}
    \caption{SegWave enables effective segmentation of tampered regions.  The input image is decomposed into frequency subbands (LL, LH, HL, HH) via the Haar Wavelet Transform. The high-frequency subbands are reweighted by an Adaptive Sub-band Attention module and recombined into an enriched representation for the image encoder. In parallel, point prompts sampled region-wise from the ground-truth mask are embedded by a prompt encoder. The image and prompt embeddings feed a transformer encoder-decoder whose mask decoder predicts the tampered regions, supervised by a weighted BCE and MSE loss. This lets SegWave fuse frequency and spatial cues for fine-grained, multi-source localization.}
    \label{fig:dwt-arch}
\end{figure*}

Yet most existing forensic models stop at exactly this binary verdict. State-of-the-art methods such as CAT-Net~\cite{cat}, SAFIRE~\cite{safire}, and DocTamper~\cite{doctamper} perform well under controlled conditions but struggle to generalize across the diversity of real manipulations such as object splicing, copy-move operations, and GAN- or diffusion-based edits~\cite{realistic,ostf}. A deeper limitation lies in what they rely on: much of their evidence comes from global frequency artifacts left behind by JPEG compression, captured through the Fast Fourier Transform (FFT)~\cite{fft,safire} or the Discrete Cosine Transform (DCT)~\cite{dct,cat,kwon2021cat}. These cues are powerful for compressed media but fade on uncompressed formats such as PNG, where no lossy footprint exists, and they offer little help in separating the multiple forgery sources that coexist within a single composite image. In high-stakes settings, where traceability and accountability matter as much as accuracy, this absence of fine-grained, source-aware localization is a serious gap.

To address this, we present \textbf{SegWave} (Segment any Manipulated Image using Wavelet Transforms), a hybrid prompt-guided framework that unifies spatial and localized frequency cues for tampering detection and localization. Rather than depending on compression artifacts, SegWave employs the Discrete Wavelet Transform (DWT)~\cite{chen2024enhancing,dwt} to capture multi-scale, spatially localized frequency variations that remain informative even on uncompressed images. An \textbf{Adaptive Subband Attention Module} then dynamically weights the most informative high-frequency subbands while suppressing structural noise, sharpening the boundaries of tampered regions. Beyond a binary verdict, SegWave integrates a point-prompt-based segmentation mechanism~\cite{sam,safire} that traces and partitions a composite image by its source regions—and crucially, it learns to do so under only weak binary-mask supervision, bridging fine-grained forensic interpretability with real-world scalability. Our key contributions are summarized as follows:

\begin{enumerate}
    \item We propose \textbf{SegWave}, a hybrid transformer framework that fuses spatial features with localized DWT-based frequency cues, exposing multi-scale anomalies that survive in uncompressed (PNG) images and sidestepping the JPEG dependence of prior baselines.

    \item We introduce an \textbf{Adaptive Subband Attention Module}  that dynamically weights the most informative high-frequency wavelet subbands, sharpening boundary localization and improving robustness to GAN- and diffusion-based edits.


    \item We conduct extensive experiments across splicing, copy-move, and generative benchmarks, showing that SegWave consistently outperforms strong baselines such as SAFIRE, CAT-Net V2, and TruFor under challenging evaluation settings.
\end{enumerate}

\noindent The rest of this paper is organized as follows: Section~\ref{sec:related} reviews related work in tampering detection. Section~\ref{sec:methodology} details the SegWave architecture. Section~\ref{sec:evaluation} presents experimental results and analysis. Finally, Section~\ref{sec:conclusion} concludes with key insights and future directions.

\section{Related Works}
\label{sec:related}

\noindent\textbf{Spatial forensic detectors.} Early methods searched for inconsistencies directly in pixel space, relying on cues such as edge and boundary anomalies~\cite{dong2022,li2023}. The shift to learned representations brought CNN-based detectors~\cite{barni2017,park2018,wang2016}, later strengthened with self-attention~\cite{liu2022} and hierarchical modeling~\cite{hao2021,wang2022} to sharpen localization. While effective on visible splices, spatial cues alone remain fragile against subtle and well-blended edits.

\noindent\textbf{Frequency-domain forensics.} To expose manipulations invisible in pixel space, a parallel line of work analyzes spectral artifacts, often fusing FFT or DCT features with spatial evidence~\cite{wang2022,cat,kwon2021cat,safire,sam,ostf}. These approaches excel on compressed media, but much of their signal originates from JPEG-specific compression footprints, which weakens their reliability on uncompressed formats.

\noindent\textbf{Multi-source partitioning.} Most detectors return a single tampered/authentic decision and cannot separate the distinct sources composited into one image. \textit{Kwon et al.}~\cite{safire} address this with point-prompt-based segmentation, recovering multiple source regions even when trained only on binary-labeled data, and offering the kind of source-aware output that high-stakes forensics demands.

\noindent\textbf{Wavelet-based forensics.} The Discrete Wavelet Transform offers a compression-agnostic alternative whose localized, multi-scale decomposition is well suited to spatially confined edits \cite{9666937,agarwal2022boosting,9207872,agarwal2023parameter}. \textit{Gadhiya et al.}~\cite{dwt-med} use a hash-based DWT scheme for medical image tampering, and \textit{Hayat and Qazi}~\cite{dwt-dct} combine DWT and DCT for copy-move detection, though both depend on handcrafted features that limit robustness to complex forgeries. Closer to our setting, \textit{Chen et al.}~\cite{chen2024enhancing} pair DWT with transformer-based attention for text tampering; motivated by this, we explore DWT for object splicing and manipulation in natural images, where its localized frequency representation can be learned end-to-end.

\noindent\textbf{Vision-language forensics.} More recent work has moved toward MLLM-based explainable localization~\cite{fakeshield,forgerygpt}, which produces interpretable, text-grounded forgery explanations but relies on large multimodal backbones and text-annotated training data. Notably, recent analysis finds that vision-language semantic priors can even degrade localization by favoring semantic plausibility over authenticity~\cite{guo2026rethinkingvlmsimageforgery}, reaffirming the value of the low-level, artifact-sensitive cues that SegWave is built around. Positioned within this landscape, SegWave instead pursues precise localization and multi-source partitioning learned purely from binary-mask supervision, similar to weakly-supervised approaches~\cite{sheng2025weaklysupervisedimageforgerylocalization}.



\section{Proposed SegWave}
\label{sec:methodology}
We introduce \textbf{SegWave} (Segment any Manipulated Image using Wavelet Transforms), a transformer-based architecture designed to effectively integrate both frequency and spatial domain features for enhanced tampering detection, which has demonstrated significant improvements across several benchmarks. 

\begin{algorithm*}[t]
\SetAlgoLined
\small
\scriptsize
\KwInput{ 
  Training images \(\{(I_k, M_k)\}_{k=1}^{K}\) where \(I_k\) is the image and \(M_k\) is the mask.\\
  Weighting factor \(\lambda\)=0.1.\\
  Positive weight \(w_+\), negative weight \(w_-\).
}
\For{each mini-batch \(\{(I_k, M_k)\}_{k=1}^{B}\)}{
    \For{each \((I_k, M_k)\) in the mini-batch}{
        1. \textbf{Wavelet Transform}: Perform single-level Haar DWT on \(I_k\) and extract \{LH, HL, HH\}.\\
        2. \textbf{Adaptive Sub-band Attention}: Compute attention scores over LH, HL, HH and obtain a weighted sum of high-frequency clues.\\
        3. \textbf{Inverse DWT}: Reconstruct enhanced image \(\hat{I}_k\) from attended high-frequency components.\\
        4. \textbf{Patch Embedding}: Apply patch embedding to both \(I_k\) and \(\hat{I}_k\) to get \(E_k\) and \(\hat{E}_k\).\\
        5. \textbf{Feature Fusion}: Tune and add \(E_k + \hat{E}_k\) to form fused representation.\\
        6. \textbf{Prompt Encoding}: Sample point prompts from \(M_k\) (one tampered, one untampered) and encode them.\\
        7. \textbf{Transformer Backbone}: Process \(E_k\) and inject adapted features from the fusion path into each block.\\
        8. \textbf{Mask Decoder}: Fuse features and predict \((\hat{Y}_k, \hat{c}_k)\).\\
        9. \textbf{Compute Losses}: 
        \[
          \mathcal{L}_{wBCE} = -\,w_+\,M_k \log(\hat{Y}_k)\;-\;w_-\,\bigl(1 - M_k\bigr)\log\bigl(1 - \hat{Y}_k\bigr)
        \]
        \[
          \mathcal{L}_{MSE} = \frac{1}{B}\sum(\hat{c}_k - c_k)^2
        \]
        \[
          \mathcal{L}_{total} = \mathcal{L}_{wBCE} + \lambda\,\mathcal{L}_{MSE}
        \]
        10. \textbf{Update Parameters}: Backpropagate \(\mathcal{L}_{total}\) and update \(\theta\).
    }
}
\caption{Proposed SegWave Training Pipeline}
\label{algocode_SegWave_dwt}
\end{algorithm*}

\subsection{Method}

A single-level Haar wavelet transform decomposes the input image into four subbands by applying low-pass ($L$) and high-pass ($H$) filtering successively along the rows and columns. This yields an approximation subband (LL) that retains the coarse image content, and three detail subbands: LH, HL, and HH, that capture high-frequency variations along the horizontal, vertical, and diagonal orientations, respectively. Since tampering artifacts such as splicing boundaries and local texture inconsistencies manifest as high-frequency discontinuities, we retain only the three detail subbands (LH, HL, HH) and discard the low-frequency approximation.

\noindent An Adaptive Sub-band Attention (ASA) module then computes dynamic attention scores for these subbands, emphasizing the most relevant high-frequency information. These weighted subbands are combined and reconstructed using inverse DWT, generating an enhanced image that accentuates high-frequency inconsistencies essential for detecting tampering. Subsequently, both the original and enhanced images pass independently through patch embedding layers. The extracted features undergo linear transformations, are tuned, and then fused into a unified representation. This joint representation is adaptively integrated into each layer of a vision transformer backbone, allowing simultaneous learning of low-level manipulation indicators and high-level semantic features.

\noindent Simultaneously, region-wise paired sampling is applied to the ground truth tampering mask, where manipulated regions are labeled as 1 and authentic regions as 0. One point from each region is sampled and encoded by a prompt encoder, embedding them in a space compatible with image embeddings. These point prompts guide the model’s attention toward relevant regions, enhancing multi-source awareness. The transformer-based mask decoder then fuses embeddings from the image and prompt encoders, leveraging learned mask and IoU tokens to generate a segmentation mask that precisely localizes the tampered regions.


\noindent Training is guided by a composite loss function: the Weighted Binary Cross Entropy loss (\(\mathcal{L}_{wBCE}\)) is used for accurate mask prediction, while the Mean Squared Error loss (\(\mathcal{L}_{MSE}\)) estimates prediction confidence. The overall objective function is defined as \(\mathcal{L}_{total} = \mathcal{L}_{wBCE} + \lambda\,\mathcal{L}_{MSE}\). The complete training pipeline is summarized in Algorithm~\ref{algocode_SegWave_dwt}. 

Weighted Binary Cross Entropy Loss:
\begin{equation}
\mathcal{L}_{wBCE} = - w_+ y \log(\hat{y}) - w_- (1 - y) \log(1 - \hat{y})
\end{equation}
where \( y \) is the probability of the ground truth label being tampered, \( \hat{y} \) is the predicted probability of a pixel being tampered, and \( w_+ \) and \( w_- \) are the weights for positive and negative samples, ensuring equal importance to both source regions. 

Mean Squared Error Loss:
\begin{equation}
\mathcal{L}_{MSE} = \frac{1}{N} \sum_{I=1}^{N} \left( \hat{c}_i - c_i \right)^2
\end{equation}
where \( c_i \) represents the ground truth confidence scores, \( \hat{c}_i \) represents the predicted confidence scores, and \( N \) is the number of samples.

The total loss for training is given by:
\begin{equation}
\mathcal{L}_{total} = \mathcal{L}_{wBCE} + \lambda \mathcal{L}_{MSE}
\end{equation}
where \( \lambda \) is the weighting factor to balance the contributions of both losses.
\subsection{SegWave$_{DCT}$ Variant}
\begin{figure}[t]
    \centering
    \includegraphics[width=0.8\linewidth]{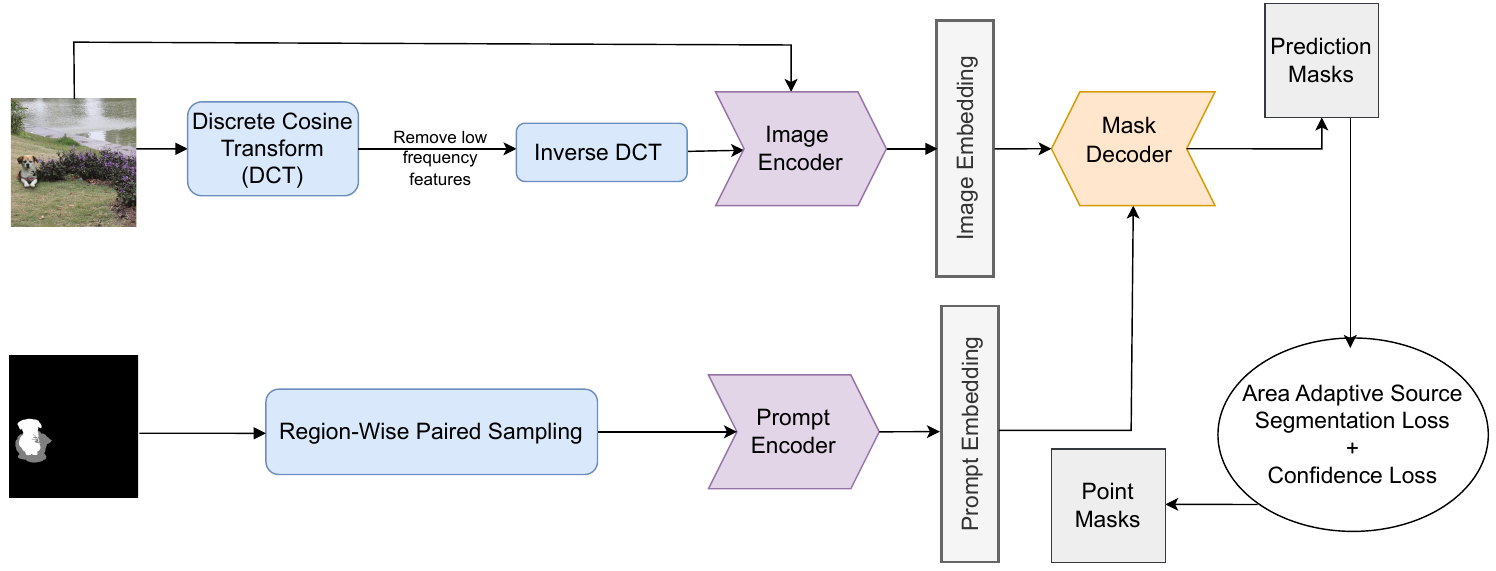}
    \caption{SegWave$_{DCT}$ as an ablation variant for effective segmentation of tampered regions. Unlike SegWave which uses DWT, this variant applies a Discrete Cosine Transform (DCT) to the input image to capture global frequency components and block-wise structural information. }
    \label{fig:dct-arch}
\end{figure}


As an ablation, we replace the wavelet decomposition with a global Discrete Cosine Transform, yielding the SegWave$_{DCT}$  variant. Because the global DCT produces a single global frequency representation rather than orientation-specific subbands, there are no subbands for the Adaptive Sub-band Attention module to weight, and it is therefore omitted in this variant. The DCT-transformed input is otherwise passed through the same encoder-decoder pipeline as SegWave, isolating the effect of the transform choice while holding the rest of the architecture fixed.

\begin{table}[t]
\centering
\small
\setlength{\tabcolsep}{4pt}
\renewcommand{\arraystretch}{1.25}
\begin{tabular}{c l ccc}
\toprule
\textbf{} & \textbf{Model} & \textbf{CocoGlide} & \textbf{Columbia} & \textbf{RealTamper} \\
\midrule
\multirow{3}{*}{\rotatebox{90}{\textbf{Prior}}}
  & \cellcolor{PriorRow} CAT-Net v2~\cite{cat}      & \cellcolor{PriorRow} 0.43 / 0.60 & \cellcolor{PriorRow} 0.86 / 0.92 & \cellcolor{PriorRow} 0.14 / 0.24 \\
  & \cellcolor{PriorRow} TruFor~\cite{guillaro2023} & \cellcolor{PriorRow} 0.52 / 0.72 & \cellcolor{PriorRow} 0.86 / 0.91 & \cellcolor{PriorRow} \textbf{0.43 / 0.53} \\
  & \cellcolor{PriorRow} SAFIRE~\cite{safire}       & \cellcolor{PriorRow} 0.63 / 0.76 & \cellcolor{PriorRow} 0.98 / 0.99 & \cellcolor{PriorRow} 0.39 / 0.50 \\
\midrule
\multirow{2}{*}{\rotatebox{90}{\textbf{Abl.}}}
  & \cellcolor{AblRow} SegWave$_{DCT}$   & \cellcolor{AblRow} 0.65 / 0.76 & \cellcolor{AblRow} 0.95 / 0.99 & \cellcolor{AblRow} \textbf{0.40} / 0.49 \\
  & \cellcolor{AblRow} SegWave w/o ASA   & \cellcolor{AblRow} 0.67 / 0.78 & \cellcolor{AblRow} 0.97 / 0.99 & \cellcolor{AblRow} 0.39 / 0.47 \\
\midrule
\rotatebox{90}{\textbf{Ours}}
  & \cellcolor{OursRow} \textbf{SegWave} & \cellcolor{OursRow} \textbf{0.69 / 0.79} & \cellcolor{OursRow} \textbf{0.99 / 0.99} & \cellcolor{OursRow} 0.36 / \textbf{0.51} \\
\bottomrule
\end{tabular}
\vspace{0.15cm}
\caption{Binary-source localization on CocoGlide~\cite{guillaro2023,coco}, Columbia~\cite{columbia}, and RealisticTampering~\cite{realistic}. Each cell reports \textbf{$F1_{fixed}$ } / \textbf{$F1_{best}$}; best per column in \textbf{bold}. Shaded green rows are ablations of our full model \textbf{SegWave}: \mbox{SegWave w/o ASA} removes the Adaptive Sub-band Attention module, while SegWave$_{DCT}$ additionally replaces the DWT with a DCT, which operates globally and therefore has no sub-bands to attend over. Epochs: SegWave$_{DCT}$ 35, SegWave w/o ASA 100, SegWave 46.}
\label{tab:safire_results}
\end{table}

\begin{table}[t]
\centering
\small
\setlength{\tabcolsep}{5pt}
\renewcommand{\arraystretch}{1.25}
\begin{tabular}{c l l cc}
\toprule
& \textbf{Dataset} & \textbf{Model} & \textbf{p\_mIoU} & \textbf{p\_ARI} \\
\midrule
\rotatebox{90}{\textbf{}}
  & \multirow{4}{*}{SAFIRE-MS-Expert-2}
    & \cellcolor{PriorRow} SAFIRE          & \cellcolor{PriorRow} 0.763 & \cellcolor{PriorRow} 0.660 \\
& & \cellcolor{AblRow} SegWave$_{DCT}$     & \cellcolor{AblRow} 0.570 & \cellcolor{AblRow} 0.514 \\
& & \cellcolor{AblRow} SegWave w/o ASA     & \cellcolor{AblRow} 0.695 & \cellcolor{AblRow} 0.630 \\
& & \cellcolor{OursRow} \textbf{SegWave}   & \cellcolor{OursRow} \textbf{0.821} & \cellcolor{OursRow} \textbf{0.710} \\
\midrule
  & \multirow{4}{*}{SAFIRE-MS-Expert-3}
    & \cellcolor{PriorRow} SAFIRE          & \cellcolor{PriorRow} 0.618 & \cellcolor{PriorRow} 0.573 \\
& & \cellcolor{AblRow} SegWave$_{DCT}$     & \cellcolor{AblRow} 0.570 & \cellcolor{AblRow} 0.514 \\
& & \cellcolor{AblRow} SegWave w/o ASA     & \cellcolor{AblRow} 0.648 & \cellcolor{AblRow} 0.637 \\
& & \cellcolor{OursRow} \textbf{SegWave}   & \cellcolor{OursRow} \textbf{0.648} & \cellcolor{OursRow} \textbf{0.641} \\
\midrule
  & \multirow{4}{*}{SAFIRE-MS-Expert-4}
    & \cellcolor{PriorRow} SAFIRE          & \cellcolor{PriorRow} 0.421 & \cellcolor{PriorRow} 0.568 \\
& & \cellcolor{AblRow} SegWave$_{DCT}$     & \cellcolor{AblRow} 0.393 & \cellcolor{AblRow} 0.513 \\
& & \cellcolor{AblRow} SegWave w/o ASA     & \cellcolor{AblRow} 0.419 & \cellcolor{AblRow} \textbf{0.587} \\
& & \cellcolor{OursRow} \textbf{SegWave}   & \cellcolor{OursRow} \textbf{0.422} & \cellcolor{OursRow} 0.577 \\
\bottomrule
\end{tabular}
\vspace{0.15cm}
\caption{Multi-source partitioning on SafireMS-Expert~\cite{safire} with 2, 3, and 4 source regions. p\_mIoU (Permuted mean IoU) and p\_ARI (Adjusted Rand Index), per \textit{Kwon et al.}~\cite{safire}. Green rows are ablations of \textbf{SegWave} (blue); best per block in \textbf{bold}.}
\label{tab:safire_multiresults}
\end{table}

\subsection{Inference with Multiple Point Prompts}

For inference, multiple-point prompts have been used to arrange in a grid pattern. Given an input image $I$ and a set of points $Q_1, \cdots, Q_N$, the model first extracts the image embedding $\mathcal{E} = E(I)$ and prompt embeddings $\mathcal{H}_i = H(Q_i)$ for all $I$. Since $\mathcal{E}$ is computed once, the process remains efficient despite multiple points. \\

\noindent Next, the mask decoder $D(\cdot, \cdot)$ generates prediction maps:
\begin{align}
(\{ Y_1, \cdots, Y_N \}, \{ c_1, \cdots, c_N \}) = D(\mathcal{E}, \{\mathcal{H}_1, \cdots, \mathcal{H}_N \}),
\end{align}
where $Y_i$ represents the predicted mask and $c_i$ its confidence score. \\

\noindent To refine predictions, we have computed representative features $\mathcal{T}_i$ by averaging image embeddings over predicted regions:
\begin{align}
\mathcal{T}_i = \frac{1}{|\mathcal{R}^{bin(\mathcal{Y}),\{1\}}|} \sum_{(m,n) \in \mathcal{R}^{bin(\mathcal{Y}),\{1\}}} \mathcal{E} [m,n].
\end{align}

\noindent  The representative features are clustered into \( M \) groups, ensuring that predictions from the same source region are grouped. From each cluster, the most confident mask is selected as the best representation of that region. For the final prediction, selected masks are merged, using softmax for multi-source segmentation and averaging for binary-source segmentation to refine the output.
\subsection{Metrics for Multi-Source Partitioning}

To evaluate the performance of multi-source partitioning, two key metrics are used: the mean Intersection over Union (mIoU) and the Adjusted Rand Index (ARI). These metrics have been utilized in SOTA, i.e., SAFIRE~\cite{safire}, and thus, adopted for comparison with SegWave and its ablation variants. The mIoU is a widely used metric in semantic segmentation that quantifies how well the model’s predicted segmentation aligns with the ground truth. SAFIRE~\cite{safire} extends the concept of permuted metrics to mIoU in multi-source partitioning, defining the permuted metric $p\_met(\cdot,\cdot)$ as follows:
\begin{align}
    p\_met(Y,X^*) = \mathop{\max}_{\tilde{X}^* \in \textnormal{Perm}(X^*)} met(Y,\tilde{X}^*),
\end{align}
where $\textnormal{Perm}(\cdot)$ denotes the set of all possible permutations along the label dimension. ARI measures the similarity between two clusters, which ranges from -1 (misclassified clusters) to 1 (perfect clustering). It handles label permutations to make it easier to evaluate label-agnostic segmentation in multi-source partitioning.

\begin{figure*}[t]
    \centering
    \includegraphics[width=0.7\linewidth]{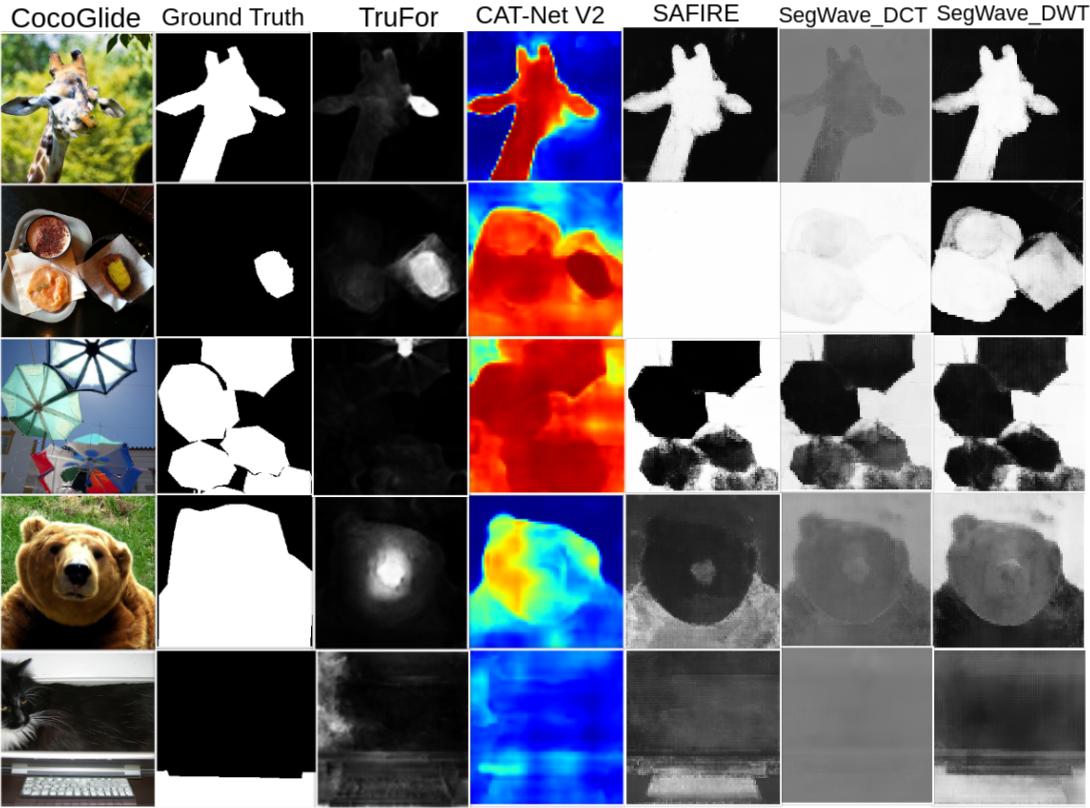}
    \vspace{-3mm}
    \caption{\small SegWave test results on CocoGlide~\cite{guillaro2023} dataset images demonstrate their superior ability to segment tampered regions compared to SegWave$_{DCT}$, SAFIRE~\cite{safire}, TruFor~\cite{guillaro2023}, and CAT-Net V2~\cite{cat}.}
    \label{fig:cocoglide-results}
    \vspace{-3mm}
\end{figure*}

\section{Implementation Details}
\label{sec:evaluation}

As defined by \textit{Guillaro et al.}~\cite{guillaro2023}, we evaluate performance using the permuted F1-Score~\cite{cat,huh2018}, considering both $F1_{fixed}$ and $F1_{best}$. $F1_{fixed}$  computes the F1-score using a threshold of 0.5 on the predicted localization map, classifying pixels with probability $\geq$ 0.5 as tampered, ensuring consistency across images. $F1_{best}$ optimizes the threshold per image to maximize the F1-score, serving as an upper-bound performance measure. We trained SegWave$_{DCT}$ and SegWave until the loss converged. During testing, we uniformly sample 256 prompt points arranged in a $16 \times 16$ grid and process them in batches of 128 to avoid GPU memory overflow during inference. The loss weighting factor 
$\lambda$ was set to 0.1, keeping the auxiliary MSE confidence term as a minor regularizer relative to the primary segmentation objective.\\

\noindent Rather than a block-based approach, we apply Global DCT and single-level Haar Wavelet Transform for SegWave$_{DCT}$ and SegWave, respectively; we assert that it is sufficient for natural image tampering, where manipulated regions tend to be relatively large compared to document text forgeries\cite{doctamper,ostf}. Our findings show that this strategy enables robust detection, making SegWave effective across diverse manipulation scenarios. The choice of Haar has been motivated by its effectiveness in prior forensic literature and its compatibility with lightweight attention mechanisms \cite{7791171,anshumaan2020wavetransform,bhattacharjee2025cae,sun2026dual}.

\subsection{Datasets}
We train SegWave on a subset of two datasets, CASIA 2.0~\cite{casia} and FantasticReality~\cite{kniaz2019point}, contributing 2{,}000 images each. We evaluate on four benchmarks held out from training: RealisticTampering~\cite{realistic} (220 images), CocoGlide~\cite{guillaro2023,coco,glide} (512 images), Columbia~\cite{columbia} (180 images), and SafireMS-Expert~\cite{safire} (238 images). SafireMS-Expert additionally provides images with 2, 3, and 4 distinct source regions, which we use for the multi-source partitioning evaluation.

\begin{figure*}[t]
    \centering
    \includegraphics[width=0.7\linewidth]{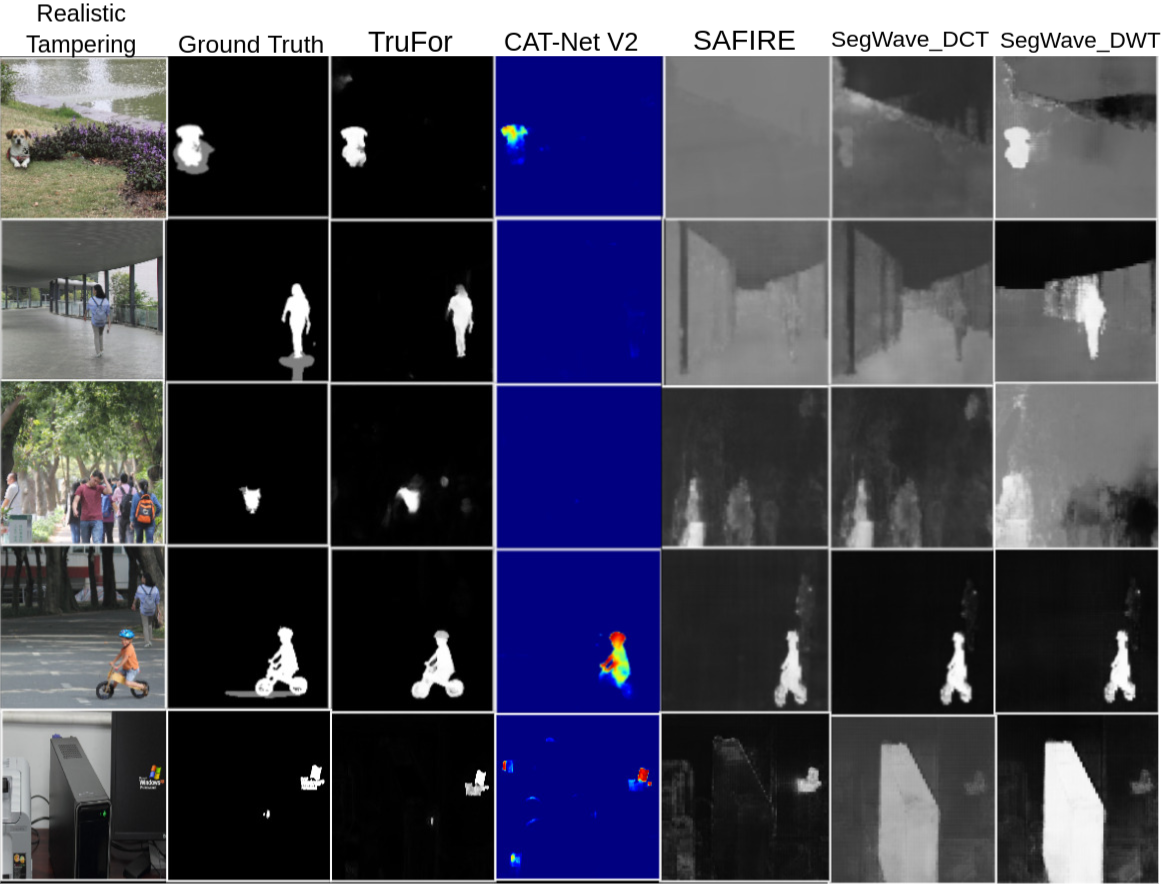}
    \vspace{-3mm}
    \caption{\small SegWave and SegWave$_{DCT}$ test results on RealisticTampering~\cite{realistic} dataset images demonstrate their superior ability to segment tampered regions compared to SAFIRE~\cite{safire} and CAT-Net V2~\cite{cat}.}
    \label{fig:realtamper-results}
\end{figure*}

\begin{figure*}[t]
    \centering
    \includegraphics[width=0.7\linewidth]{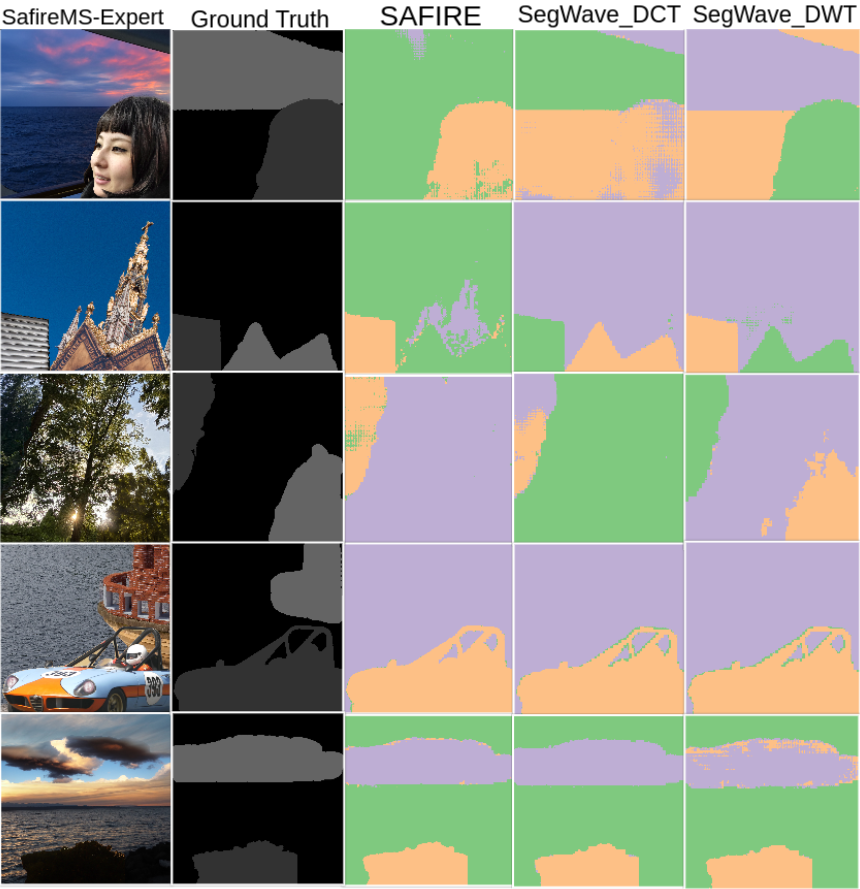}
    \vspace{-3mm}
    \caption{\small SegWave outperforms all other models on the SafireMS-Expert dataset~\cite{safire}, achieving the best performance across all evaluation metrics.}
    \label{fig:ms-all-results}
    \vspace{-3mm}
\end{figure*}

\section{Experimental Results \& Analysis}

\noindent\textbf{Binary-source localization.} As shown in Table~\ref{tab:safire_results}, SegWave achieves the strongest results on CocoGlide and Columbia. On CocoGlide, it improves $F1_{\text{fixed}}$ from 0.43 (CAT-Net~v2) and 0.52 (TruFor) to 0.69 - gains of 0.26 and 0.17, and improves on SAFIRE's 0.63 by 0.06, with a similar ordering on $F1_{\text{best}}$. On Columbia, SegWave reaches near-saturated performance, matching or exceeding all baselines. SegWave does not win everywhere: on RealisticTampering it trails TruFor's 0.43 by 0.07 on $F1_{\text{fixed}}$, a gap we analyze below. \\

\noindent\textbf{The role of Adaptive Sub-band Attention (ASA).} The ablation rows isolate the contribution of the attention module. Removing it (SegWave w/o ASA) leaves the DWT-enhanced input intact but weights all high-frequency subbands equally, while replacing the transform entirely (SegWave$_{DCT}$) removes the notion of orientation-specific subbands altogether. On binary localization, the gain from ASA is modest ($0.67 \rightarrow 0.69$ $F1_{\text{fixed}}$ on CocoGlide), but its value becomes pronounced on the harder multi-source task (Table~\ref{tab:safire_multiresults}): on SafireMS-Expert-2, ASA lifts p\_mIoU from 0.695 to 0.821 and p\_ARI from 0.630 to 0.710 over the equally-weighted variant. This matches the intuition behind the module: when multiple manipulated sources coexist, their boundary signatures reside in different orientation subbands, and adaptively reweighting LH, HL, and HH lets the model attend to the discriminative subband per source rather than diluting all three. SegWave$_{DCT}$, lacking subbands to reweight, trails the full model across every multi-source setting, confirming that the gain stems from oriented, adaptively weighted decomposition rather than from frequency information alone.

\noindent\textbf{When DCT helps.} The one setting where SegWave$_{DCT}$ is competitive is RealisticTampering, where it edges out SegWave on $F1_{\text{fixed}}$ (0.40 vs 0.36). We attribute this to the dataset: although stored as TIFF, these images have likely undergone prior lossy JPEG compression, leaving DCT-aligned periodic artifacts that a DCT model captures directly. This is consistent with the broader pattern we observe: global DCT is better matched to compression-induced cues, whereas oriented DWT with ASA is better suited to spatially confined boundary inconsistencies of splicing and generative edits.  \\

\noindent\textbf{Multi-source partitioning.} Table~\ref{tab:safire_multiresults} compares SAFIRE, SegWave$_{DCT}$, SegWave w/o ASA, and the full SegWave across 2-, 3-, and 4-source settings. SegWave attains the best p\_mIoU in every setting and the best p\_ARI on the 2- and 3-source splits, with its clearest advantage on the 2-source case (0.821 p\_mIoU vs SAFIRE's 0.763). As the number of sources grows to four, all methods converge and the margins narrow, reflecting the increasing difficulty of separating many co-located manipulations. Qualitatively, Figures~\ref{fig:cocoglide-results}--\ref{fig:ms-all-results} show that SegWave produces sharper, better-separated source regions than SAFIRE, TruFor, and CAT-Net~v2, consistent with the quantitative trends.

\section{Conclusion}
\label{sec:conclusion}

We presented SegWave, a prompt-guided framework that combines a Haar wavelet decomposition with an Adaptive Sub-band Attention module for image tampering localization and multi-source partitioning. By operating on oriented, spatially localized frequency subbands rather than global-frequency artifacts, SegWave remains effective on uncompressed formats where compression-dependent cues fade. It effectively localizes the spatially-confined boundary inconsistencies characteristic of splicing and generative edits. Our ablations isolate the source of these gains: rather than weighting the LH, HL, and HH subbands equally, adaptively reweighting them prevents informative cues from being diluted. This effect is most pronounced on the harder multi-source task, where SegWave clearly improves over both the equally-weighted variant and the DCT variant. The comparison between the two transforms also reveals a complementary structure: global DCT better captures compression-induced artifacts, while oriented DWT better captures localized boundary cues. This suggests a natural direction for future work: unifying both transforms so that a single model can draw on the compression-awareness of DCT and the localized sensitivity of wavelets, potentially yielding robust and generalizable tampering detection.

\section*{Acknowledgements}
V. Kumar is partially supported through the Visvesvaraya PhD Fellowship, implemented by the Ministry of Electronics and Information Technology (MeitY), Government of India. 

\bibliographystyle{splncs04}
\let\oldthebibliography\thebibliography
\renewcommand{\thebibliography}[1]{\oldthebibliography{99}}
\bibliography{egbib}
\end{document}